\documentclass{article}

     \usepackage[preprint]{neurips_2023}

\usepackage[utf8]{inputenc} % allow utf-8 input
\usepackage[T1]{fontenc}    % use 8-bit T1 fonts
\usepackage{hyperref}       % hyperlinks
\usepackage{url}            % simple URL typesetting
\usepackage{booktabs}       % professional-quality tables
\usepackage{amsfonts}       % blackboard math symbols
\usepackage{nicefrac}       % compact symbols for 1/2, etc.
\usepackage{microtype}      % microtypography
\usepackage{xcolor}         % colors
\usepackage{graphicx}
\usepackage{amsmath}%wumin

\title{PruneSymNet: A Symbolic Neural Network and Pruning Algorithm for Symbolic Regression}

\author{
  Min~Wu \\
  \And
  Weijun~Li \thanks{Corresponding author: Weijun~Li. \\Min~Wu, Weijun~Li, Lina~Yu, Linjun Sun, Jingyi~Liu,
  	and Wenqiang~Li are with AnnLab, Institute of Semiconductors, Chinese
  	Academy of Sciences, Beijing 100083, China, Center of Materials
  	Science and Optoelectronics Engineering \& School of Microelectronics,
  	University of Chinese Academy of Sciences, Beijing, 100049, China
  	and Beijing Key Laboratory of Semiconductor Neural Network Intelligent
  	Sensing and Computing Technology, Beijing 100083, China. (e-mail:wumin; wjli; yulina; liwenqiang; liujingyi; liyanjie; mlhao@semi.ac.cn; ). }\\
  \And
  Lina~Yu\\
  \And
  Wenqiang~Li\\
  \And
  Jingyi~Liu\\
 \And
 Yanjie~Li\\
  \And
  Meilan~Hao\\
}

\begin{document}

\maketitle

\begin{abstract}
  Symbolic regression aims to derive interpretable symbolic expressions from data in order to better understand and interpret data. %which plays an important role in knowledge discovery and interpretable machine learning. 
  In this study, a symbolic network called PruneSymNet is proposed for symbolic regression. This is a novel neural network whose activation function consists of common elementary functions and operators. The whole network is differentiable and can be trained by gradient descent method. Each subnetwork in the network corresponds to an expression, and our goal is to extract such subnetworks to get the desired symbolic expression.
  Therefore, a greedy pruning algorithm is proposed to prune the network into a subnetwork while ensuring the accuracy of data fitting. The proposed greedy pruning algorithm preserves the edge with the least loss in each pruning, but greedy algorithm often can not get the optimal solution. In order to alleviate this problem, we combine beam search during pruning to obtain multiple candidate expressions each time, and finally select the expression with the smallest loss as the final result. It was tested on the public data set and compared with the current popular algorithms. The results showed that the proposed algorithm had better accuracy.
\end{abstract}

\section{Introduction}
Symbolic regression is the process of finding the mathematical expression describing the system given the input and output of an unknown system. It has important application in knowledge discovery and explicable machine learning, and has important theoretical significance and practical value.

As we know, deep neural network is in essence a huge mathematical expression with strong fitting ability. However,  it is an unexplainable black box, since the inside of neural network is too complex to be described by a simple expression.
If we can simplify the neural network so that it can be described by a simple mathematical expression while maintaining its fitting ability, we will get an interpretable regression model.
Inspired by this idea, researchers proposed symbolic neural network for symbolic regression. A typical example is EQL network \cite{eql,Kim}, which replaces the activation function of feedforward neural network with common elementary functions and operators, and combines sparse optimization in parameter learning, so as to obtain a sparse subnetwork and a symbolic expression.

However, it is difficult to obtain sufficiently simple expressions through sparse optimization. The EQL network still contains a large number of non-zero parameters after sparse optimization, which makes the obtained expressions too complex and thus reduces the interpretability. Moreover, elementary functions and operators, especially division operators, are not suitable for activation functions, because it is easy to cause gradient explosion and unstable training. For this reason, EQL network only contains division operators in the last layer, which leads to the deficiency of EQL representation ability.

In order to solve these problems, a symbolic neural network PruneSymNet has been proposed in this study, which has strong representation ability. In order to avoid the difficulty of sparse optimization, we propose a greedy pruning algorithm inspired by model pruning algorithm\cite{pruneDNN} to obtain a sufficiently simple subnetworks, so as to obtain an expression with lower complexity. In addition, we solved the gradient explosion problem in training PruneSymNet, so that the training is stable even when every layer contains division operators.

The main contributions of this study are summarized as follows:

\noindent 1. A symbolic network PruneSymNet is proposed which can represent any expression. The network is differentiable and can be trained by gradient descent method.

\noindent 2. A pruning method is proposed to obtain a sufficiently simple subnetworks, so as to obtain an expression with lower complexity.

\noindent 3. An improved gradient descent method is proposed to avoid gradient explosion, and division operators can be included in every layer of PruneSymNet.

\noindent 4. A coefficient post-processing method is proposed for the original expression obtained by pruning, so that a more concise expression can be obtained.

The remainder of this paper is organized as follows. Section \ref{Sec:relatedwork} presents a review of the existing methods on symbolic regression. Section \ref{Sec:SymNet} describes the design of PruneSymNet. Section \ref{sec:alg} introduces the algorithm of solving prunSymNet network, including gradient descent module, greedy pruning module and coefficient post-processing module.
Section \ref{Sec:Experiments} compares proposed algorithm with that of state-of-the-art algorithms. Section \ref{Sec:discussion} discussed some characteristics of the algorithm, including the contributions of each module on the performance, the equivalent subnetwork in pruning and the limitations of the proposed algorithm.  Finally, the conclusion of the study are presented in Section \ref{sec:conclusion}.

\section{Related works}\label{Sec:relatedwork}
Symbolic regression has been studied for a long time, and the process of Kepler's discovery of planetary orbit equation is a typical example. 
%The difficulty of symbolic regression is that its solution space grows exponentially with the increase of expression length, which is a NP hard combinatorial optimization problem. 
Symbolic regression methods can be roughly divided into search-based methods\cite{gp,dsr,eql} and supervised-based pretraining methods\cite{scales, SymbolicGPT, e-e}. 
%Search-based methods search in the solution space of symbolic expressions to get expressions that meet the conditions. The supervised pre-training based methods accumulate solving experience through offline learning of a large number of training samples. When facing new problems, solving experience is utilized to achieve the purpose of solving problems quickly. 
%The advantages of search-based method are flexible and universal, but the disadvantages are low efficiency and time consuming.
The supervision-based method has the advantage of being fast speed, but it is inflexible and the performance depends on training data.
%The supervised learning based method has the advantage of fast speed but poor flexibility, and its performance largely depends on training samples. 
For problems where the number of variables and the range of variables are different from training samples, the performance is poor.
%Both the search-based method and the supervision-based pre-training method have their own advantages and disadvantages. Different methods should be adopted in application according to specific problems. 
So this study focuses on search-based method, although it is a bit time-consuming. The commonly used algorithms include genetic algorithm\cite{gp,gp2}, EQL algorithm\cite{eql}, DSR algorithm\cite{dsr}, AIFeynman method\cite{aifeyman,aifeyman2}, etc., which are introduced below.

Genetic algorithm\cite{gp,gp2} is the most classic symbolic regression algorithm. In genetic algorithm, the expression is expressed in the form of binary tree(symbol tree), and the new generation of expression individuals is generated by a series of genetic operations such as mutation and crossover. Each individual performance is evaluated by fitness function, and the individuals with good performance are retained for the next generation of evolution. In the process of evolution, more accurate expressions are generated. %, typical methods include MRGP[15], GSGP[16], FFX[17], etc.
 However, the expression obtained by genetic algorithm is often too complicated in order to pursue precision.

The EQL\cite{eql} algorithm replaces activation functions in feedforward neural networks with common elementary functions and operators. A sparse network is obtained by gradient descent and sparse optimization, and then the sparse network is restored to a mathematical expression.
However, sparse optimization is often difficult. The sparse EQL network still contains a large number of non-zero parameters, resulting in overly complex expressions and loss of interpretability.

DSR\cite{dsr} is based on reinforcement learning, which represents symbol tree as symbol sequence through first order traversal, and searches the symbol sequence that meets the requirement through strategy gradient method in reinforcement learning.
%The idea of DSR is to make the symbol sequence with small expression errors appear more and more likely.
DSR performs well on many data sets, but the expression input/output data is only used to calculate the reward and not used in the policy network, which limits its further performance improvement.

AIFeynman\cite{aifeyman,aifeyman2} is a symbolic regression method for physical formulas. It takes advantage of dimensional restrictions, variable separability, symmetry and other properties in physical laws to decompose problems into small modules, thus reducing the complexity of problems.
One key for AIFeynman is to train a neural network in advance to fit the input and output of the current problem, and the gradient information of the trained neural network is used to approximate the gradient of the expression to be solved. AIFeynman is not a general purpose symbolic regression algorithm, and it is not reasonable to use a neural network to fit input and output data, because there may not be sufficient data in the actual situation to train the neural network.

For more symbolic regression algorithms, refer to \cite{zongshu}.

\section{PruneSymNet Network structure}\label{Sec:SymNet}
\begin{figure}[!htbp]
	\centering
	\includegraphics[width=0.5\textwidth]{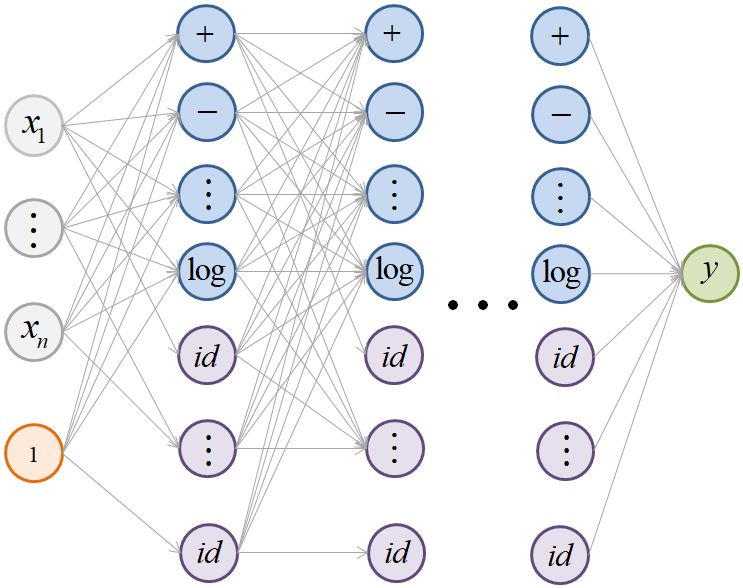}
	\caption{Schematic diagram of PruneSymNet structure.} \label{PruneFullNet}
\end{figure}

Figure \ref{PruneFullNet} shows the overall framework of DeepSymNet. The first layer of PruneSymNet is the input layer, the middle  is the hidden layer, and the last is the output layer.
The nodes of the hidden layer consist of elementary functions and operators $(+, -, \times, \div, sin, cos, exp, log, square, id)$. Note that $id$ is the identity operator($id(x)=x$), which is the same as defined in EQL\cite{eql}.

The node with operator $id$ has a one-to-one correspondence with the nodes of the previous layer. Its function is to enable each layer to make use of all the information of the previous layers. Therefore, the number of $id$ operators in each layer is equal to the number of nodes in the previous layer. 
The other operators(called ordinary operators) appear only once per hidden layer and are fully connected to the previous layer.
Each ordinary operator has one or two full connections to the previous layer, depending on whether it is a unary or binary operator.
If the number of nodes in layer $i$ is $n_i$ and the number of operators except $id$ is $m$, then the number of nodes in the hidden layer meets $n_{i+1}=n_i+m$.

The connection between the $id$ operator and the previous layer is fixed, and the edge has no weight to learn. The weights on the full connection edges of ordinary operators need to be learned.

Note that the input layer has a special node, "1", whose input is always constant 1, which is equivalent to turning the input data into homogeneous coordinates. The weight of the edge connected to node "1" actually becomes the bias term $b$ in the linear map $WX+b$. Therefore, linear mapping in PruneSymNet does not require a separate bias term $b$, which makes pruning easier.

\begin{figure}[!htbp]
	\centering
	\includegraphics[width=0.40\textwidth]{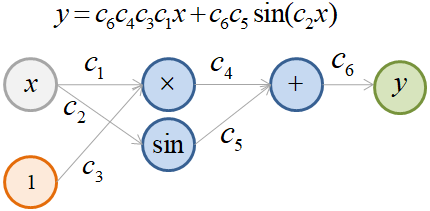}
	\caption{A sub-network represents a symbolic expression in PruneSymNet. (Take $y = {c_6}{c_4}{c_3}{c_1}x + {c_6}{c_5}\sin ({c_2}x)$ as an example).} \label{SubNet1}
\end{figure}

 Each subnetwork represents an expression in PruneSymNet. If the linear mapping corresponding to each node in a subnetwork preserves only one edge, we call such a subnetwork a minimalist network, as shown in Figure \ref{SubNet1}. It is easy to verify that PruneSymNet can represent any symbolic expression if it has sufficient layers.

\section{PruneSymNet Network Solving Algorithm}\label{sec:alg}
The PruneSymNet network solving algorithm consists of a gradient descent algorithm module, a greedy pruning module, and a coefficient post-processing module.
\subsection{Gradient descent algorithm for PruneSymNet}
The purpose of PruneSymNet parameter learning is to fit sample points $ \{({X_i}, {y_i}) | i=1, N \} $, so as to minimize the error between the network output $net(X_i)$ and $y_i$. The mean square error is used as the loss function, as shown in the formula
(\ref{equ:obj}), where $||W|{|_{0.5}}$ is the $L_{0.5}$ norm\cite{Kim}, making the network sparse and beneficial for pruning.
\begin{equation}\label{equ:obj}
	loss = \frac{1}{n}\sum\limits_{i = 1}^n {{{({y_i} - net({X_i}))}^2}}+||W|{|_{0.5}}
\end{equation}

The PruneSymNet network uses gradient descent method for parameter learning. In fact, elementary functions and operators are not suitable activation functions, since Nan values and gradient explosions tend to occur in the training process, leading to unstable network training and even crash. We found that the main causes of Nan value and gradient explosion are multiplication operator $\times$, division operator $\div$, exponential function $exp$. To solve this problem, we add protection to these activation functions.
Specifically, if the output $v$ of these activation functions exceeds a certain threshold value $Th $, the output will be multiplied by a certain coefficient ($v= \lambda \times v $) so that the value of $v $ does not exceed $Th$.
The gradient truncation technique is also used to limit the maximum gradient. After adopting these protections, the network training is stable.

\subsection{Greedy Pruning Algorithm Module}
\begin{figure}[!htbp]
	\centering
	\includegraphics[width=0.50\textwidth]{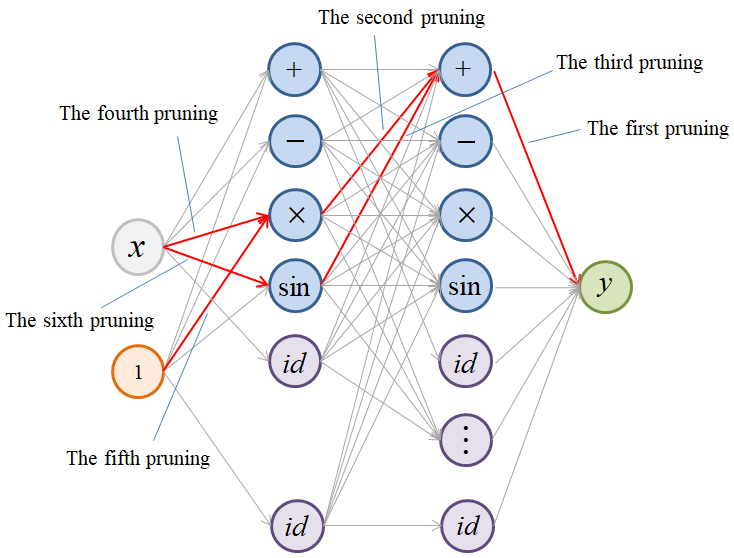}
	\caption{Schematic diagram of greedy pruning.} \label{pruneProcess}
\end{figure}

The PruneSymNet network undergoes pruning every certain number of training epochs. The proposed pruning algorithm is a greedy algorithm. The pruning starts from the network output node and proceeds from the back to the front. 
%When pruning each node, 
The edge with the minimum loss is retained in each group of full connection edges, and the weight of the other edges is set to 0 to complete the pruning of the current node.
So the node with binary operator retain two edges, and the node with unary operator retain one edge after pruning. It can be seen that after pruning, a minimalist subnetwork is ultimately obtained, resulting in a relatively simple expression. For example, the pruning process of the sub network shown in Figure \ref{SubNet1} is shown in Figure \ref {pruneProcess}.

Greedy pruning may not yield the subnetwork with the least loss. To alleviate this problem, we introduce beam search during the pruning process. Set the Beam Size to $B$, then each pruning preserves $B$ candidate edges with the lowest loss, resulting in $B$ candidate expressions. Select the one with the smallest error from the $B$ candidate expressions as the final expression.

We also found that, the network often falls into a local optimal solution after gradient descent, when the pruning results do not change. If we consider the pruning process as reinforcement learning, the pruning strategy in this study is a deterministic strategy, and the local optimal solution is caused by the insufficient exploration rate of the deterministic strategy. To alleviate this problem we propose two improvement methods.

One method is to use softmax\cite{nndl} to calculate the probability according to the loss corresponding to each edge, and carry out pruning according to the probability. In order to balance exploration and utilization, only the first two layers of the network are pruned according to the probability, and the other layers still adopt greedy pruning strategy.

Another method is to increase the learning rate by a certain value if more than 80\% of the candidates are the same as those of the last pruning results, and reset the learning rate to the original value until the overlap rate of the two pruning results is less than 80\%.
%Otherwise, continue to increase the learning rate. 
The principle is that increasing the learning rate can jump out of the local optimal solution. 

Note that after pruning, the weight will be restored to the value before pruning.

\subsection{Expression coefficient post-processing module}
The coefficients of the expression obtained by pruning are not accurate, and many coefficients can be merged. For example, the expression $y={w_6} {w_4} {w_3} {w_1} x+{w_6} {w_5} sin ({w_2} x)$ and
$y={c_1} x+{c_2} sin ({c_3} x)$ is equivalent. 
More importantly, expression sometimes contain redundant items that need to be removed.
For example, for solving the expression (\ref{equ:objExpress}),
\begin{equation}\label{equ:objExpress}
y=\frac{{{x_1}^3}}{5} - {x_1} + \frac{{{x_2}^3}}{2} - {x_2}
\end{equation}
the following expression is obtained after pruning.
\begin{equation}\label{equ:orgExpress}
	\begin{gathered}
		y={c_3}{x_1}^4 + {c_3}{x_1}^3{x_2} + {\color{red}{c_7}{x_1}^3} + {c_{13}}{x_1}^2{x_2}^2 + {c_{17}}{x_1}^2{x_2} + {c_4}{x_1}^2 + {c_{15}}{x_1}{x_2}^3 +  \hfill \\
		{c_{12}}{x_1}{x_2}^2 + {c_{16}}{x_1}{x_2} + {\color{red}{c_2}{x_1}} + {c_6}{x_2}^4 + {\color{red}{c_{10}}{x_2}^3} + {c_5}{x_2}^2 + {\color{red}{c_9}{x_2}} + {c_{11}}\log {({x_2})^2} \hfill \\
		+ {c_8}\log ({x_2}) + {c_1} \hfill \\ 
	\end{gathered} 
\end{equation}
Where, the red part needs to be retained, while the other parts are redundant items.

\begin{figure}[!htbp]
	\centering
	\includegraphics[width=1.02\textwidth]{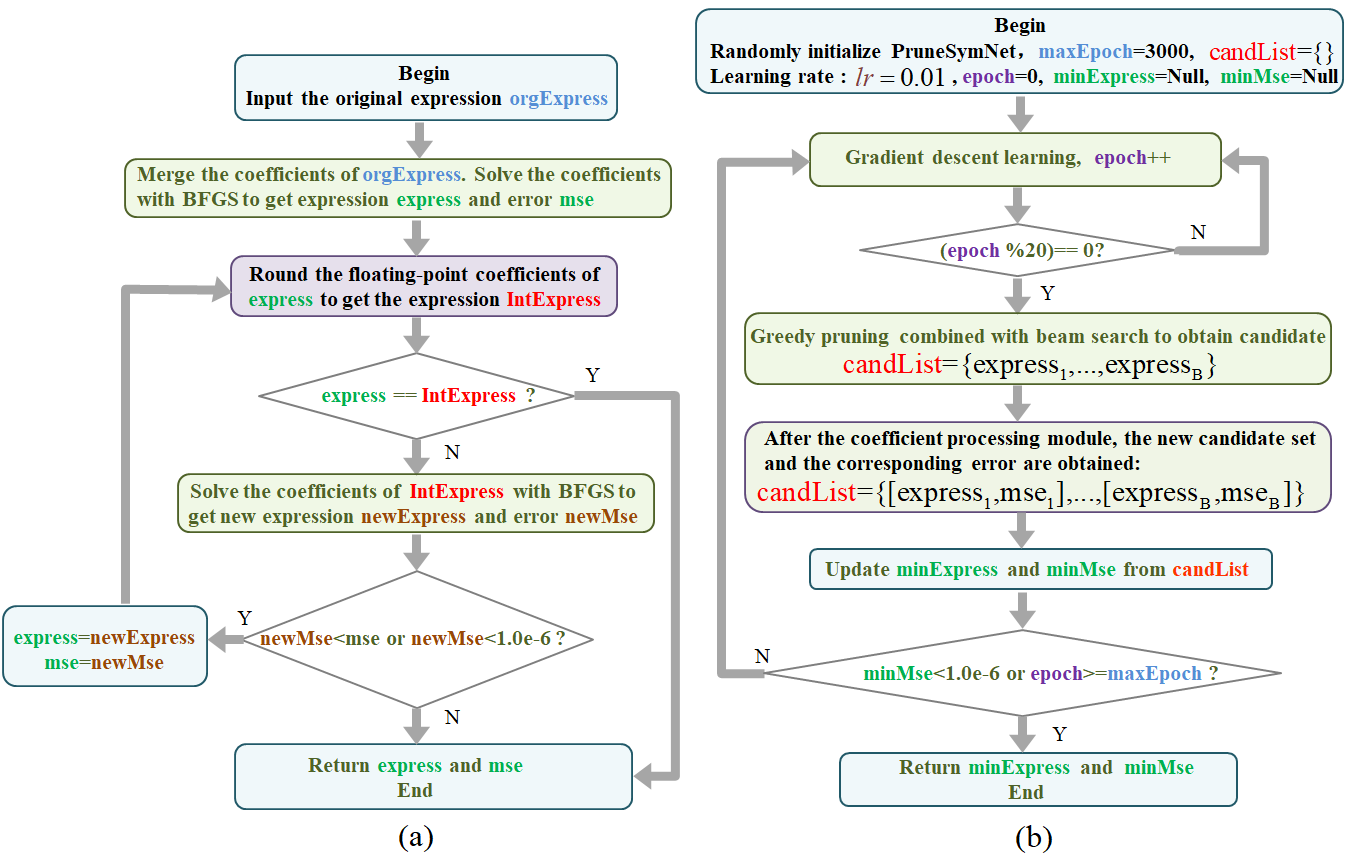}
	\caption{(a):Flow chart of coefficient post-processing; (b):Overall flowchart of the proposed algorithm.} \label{tail-all}
\end{figure}

In order to obtain more accurate coefficients and remove redundant terms, we propose a post-processing algorithm. Firstly, construct the objective function shown in (\ref{equ:objc1}) for the original expression obtained by pruning, and solve it using the BFGS \cite {nndl, NumericalOptimization} algorithm, where $C $ represents the set of all coefficients.
\begin{equation}\label{equ:objc1}
	\mathop {\min }\limits_C \sum\limits_{i = 1}^n {{{\left( {{y_{\text{i}}} - f({X_i},C)} \right)}^2}}.
\end{equation}
After solving by BFGS, the coefficients of certain redundant terms are generally close to 0, and for some integer coefficients of original expressions, the floating-point coefficients obtained will also be close to the integers. Therefore, if the error between a floating-point coefficient and its closest integer is less than a certain threshold, the coefficient will be rounded to the nearest integer, and then BFGS will be used to solve other floating-point coefficients again. Repeat the iteration until no coefficients meet the rounding conditions or the error does not decrease. The flowchart is shown in Figure\ref{tail-all} (a).
\iffalse
\begin{figure}[!htbp]
	\centering
	\includegraphics[width=0.70\textwidth]{processTail.png}
	\caption{Flow chart of coefficient post-processing.} \label{processTail}
\end{figure}
\fi
For the expression (\ref{equ:orgExpress}), the resulting expression is
\[y=0.19999999267510496{x_1}^3 - {\text{ }}{x_1}{\text{ }} + {\text{ }}0.49999999889887176{x_2}^3{\text{ }} - {\text{ }}{x_2}\]

During the pruning process, the expression $minExpress$ with the smallest error $minMse$ obtained so far are saved as the final result. The whole algorithm flow chart is shown in Figure\ref{tail-all} (b).
\iffalse
\begin{figure}[!htbp]
	\centering
	\includegraphics[width=0.80\textwidth]{pruneAll.png}
	\caption{Overall flowchart of pruning algorithm.} \label{pruneAll}
\end{figure}
\fi

\section{Experimental results}\label{Sec:Experiments}
\subsection {Experimental parameter settings}
In the experiment, PruneSymNet had six hidden layers and the number of input data points is 128 for each test expression. The sampling interval of the variable is $[-5,5] $. The Adam optimizer is used for parameter learning, with a maximum number of iterations of $maxEpoch=3000$. The beam size $B$ for beam search is set to 5. The algorithm is implemented using PyTorch.

\subsection {Introduction to Test Datasets}
Six test data sets were used in this study: Koza\cite{gen}, Korns\cite{gen}, ODE\cite{zongshu}, Livermore\cite{dso},
AIFeynman2\cite{aifeyman} and AIFeynman3\cite{aifeyman}, where AIFeynman2 and AIFeynman3 both come from data set AIFeynman\cite{aifeyman}. The number of expression variables in AIFeynman2 is less than or equal to 2, and in AIFeynman3 is equal to 3. Table \ref{L2_1} lists the number of expressions contained in each dataset. The detailed information on theses dataset can be found in the supplementary materials.
\begin{table}[!htbp]
	\centering
	\caption{The number of expressions contained in each dataset}\label{L2_1}
	\begin{tabular}{cccccc}
		\hline
		Koza&Korns&ODE&Livermore&AIFeynman2&AIFeynman3\\
		\hline
		16 &7&12&16&12&24\\
		\hline
	\end{tabular}
\end{table}

\subsection{Test results}
The algorithms we used for comparison are EQL\cite{eql}, GP\cite{gp}, DSR\cite{dsr}. All algorithms were independently tested three times on each dataset, and the best of the three results was selected for comparison. 

\subsubsection{Comparison of MSE}
\begin{figure}[!htbp]
	\centering
	\includegraphics[width=1.0\textwidth]{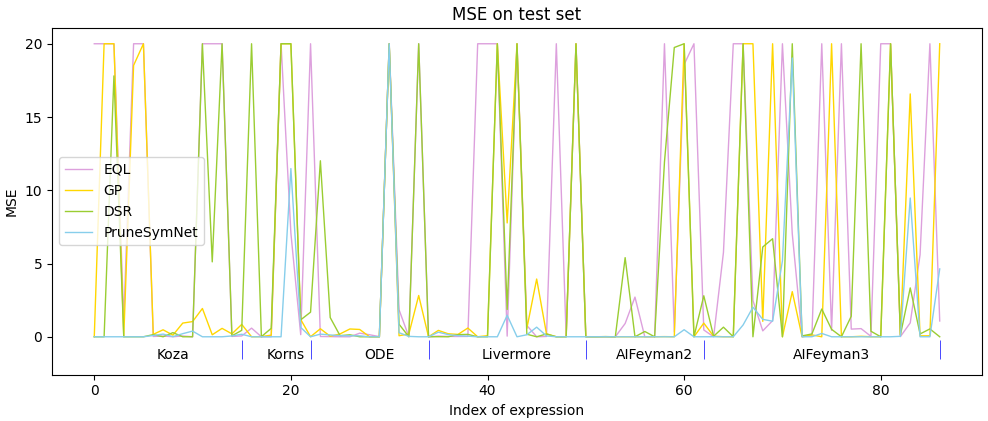}
	\caption{The $MSE$ on test set.} \label{mse-pub}
\end{figure}

The performance evaluation metric used in this section is the mean square error (MSE) of the expression.
Figure \ref{mse-pub} shows the $MSE$ graph for each algorithm on each dataset.  It can be seen from figure \ref{mse-pub} that the proposed algorithm has the smallest $MSE$, indicating that PruneSymNet has the highest accuracy.
In addition, we also compared the number of votes (the number of winning expressions) between PruneSymNet and the comparison algorithm on each dataset in order to compare the performance more clearly. The specific voting method is that for a certain expression, if the $MSE$ of PruneSymNet is less than or equal to the comparison algorithm, the number of votes on PruneSymNet increases by one, and vice versa, the number of votes on the comparison algorithm increases by one.
The voting results are shown in Table\ref{L2_2}, from which it can be seen that the accuracy of PruneSymNet is better than that of the compared algorithms.
\begin{table}[!htbp]
	\centering
	\caption{Comparison of the number of votes between PruneSymNet and the comparison algorithm on the test dataset }\label{L2_2}
	\begin{tabular}{cccccc}
		\hline
		Data Sets&EQL&GP&DSR\\
		\hline
		Koza &\bf 9:7 &\bf 15:1 &\bf 12:4  \\
		\hline
		Korns &\bf 5:2 &\bf 7:0 &\bf 7:0 \\
		\hline
		ODE &\bf 7:5 &\bf 9:3 &\bf 10:2  \\
		\hline
		Livermore &\bf 9:7 &\bf 16:0 &\bf 13:4  \\
		\hline
	    AIFeynman2 &\bf 11:1 &\bf 10:4 &\bf 11:3  \\
	    \hline
	    AIFeynman3 & \bf 18:6 & \bf 19:6 &\bf 20:4 \\
	    \hline
	\end{tabular}
\end{table}
\subsubsection{Comparison of the number of optimal solutions}
For a test sample, if the expression solved by the symbolic regression algorithm is the same as the true expression corresponding to the sample, it is said to have obtained the optimal solution, which is a challenging task. Table \ref{L2_real} lists the comparison of the number of optimal solutions obtained by the algorithms tested on six datasets. The results show that PruneSymNet obtains the most optimal solutions, further verifying the effectiveness of the proposed algorithm. The detailed information on the optimal solution obtained by each algorithm can be found in the supplementary materials.
\begin{table}[!htbp]
	\centering
	\caption{Comparison of the number of optimal solutions on the test dataset}\label{L2_real}
	\begin{tabular}{ccccc}
		\hline
		Data Sets&EQL&GP&DSR&PruneSymNet\\
		\hline
		Koza & 0 &  1 &  8 & \bf 9 \\
		\hline
		Korns & 0 &  0 &  0 & \bf 5 \\
		\hline
		ODE & 0 &  2 &  3 & \bf 4 \\
		\hline
		Livermore & 0 &  2 &  6 & \bf 8 \\
		\hline
		AIFeynman2 & 0 &  5 &  4 &  \bf 9 \\
		\hline
		AIFeynman3 & 0 &  6 &  6 & \bf 10 \\
		\hline
	\end{tabular}
\end{table}
\section{Discussion}\label{Sec:discussion}
\subsection{Ablation experiment to explore the contribution of the algorithm modules on performance}
Gradient descent and greedy pruning are the most important modules in the proposed algorithm. To explore their contribution on algorithm performance, we designed ablation experiments. We constructed ModelNoGd and ModelRandPrune based on the complete algorithm, where ModelNoGd removed the gradient descent module and the pruning algorithm in ModelRandPrune was replaced by random pruning. We compared the performance of the complete model and these two models, with the indicators of the number of votes and optimal solutions. Each model iterated up to 1000 times, and the results are shown in Tables \ref{La_1}-\ref{La2_real}. It can be seen from the results that the accuracy of the complete model is much higher than that of ModelNoGd and ModelRandPrune,
%algorithm performance has significantly decreased after removing these two modules, 
which proves the importance of gradient descent and greedy pruning in the proposed algorithm.

\begin{table}[!htbp]
	\centering
	\caption{Comparison of the number of votes between PruneSymNet and ModelNoGd, ModelRandPrune}\label{La_1}
	\begin{tabular}{ccccc}
		\hline
		Data Sets&ModelNoGd&ModelRandPrune\\
		\hline
		Koza &\bf 12:4 &\bf 11:6  \\
		\hline
		Korns &\bf 7:2 &\bf 6:3 \\
		\hline
		ODE &\bf 8:4 &\bf 8:4  \\
		\hline
		Livermore &\bf 13:5 &\bf 9:8  \\
		\hline
		AIFeynman2 &\bf 8:5 &\bf 8:5  \\
		\hline
		AIFeynman3 &\bf 20:4 &\bf 16:9 \\
		\hline
	\end{tabular}
\end{table}

\begin{table}[!htbp]
	\centering
	\caption{Comparison of the number of optimal solutions between PruneSymNet and ModelNoGd, ModelRandPrune}\label{La2_real}
	\begin{tabular}{ccccc}
		\hline
		Data Sets&ModelNoGd&ModelRandPrune&PruneSymNet(Complete Model)\\
		\hline
		Koza & 0 &  1 &\bf  8 \\
		\hline
		Korns & \bf 3 &\bf  3 &\bf 3 \\
		\hline
		ODE & 1 &  1 & \bf 2 \\
		\hline
		Livermore & 2 &  2 &\bf  4 \\
		\hline
		AIFeynman2 & 5 &  5 & \bf 6 \\
		\hline
		AIFeynman3 & 0 &  1 & \bf 3 \\
		\hline
	\end{tabular}
\end{table}

\subsection{Equivalent subnetwork}
\begin{figure}[!htbp]
	\centering
	\includegraphics[width=0.8\textwidth]{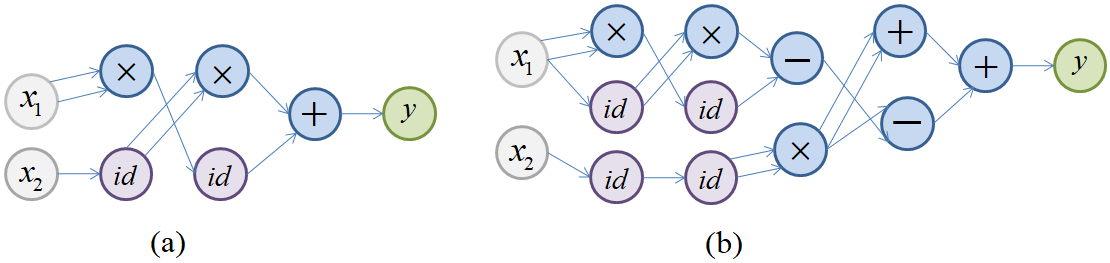}
	\caption{Equivalent subnetwork(Take $y={x_1}^2+{x_2}^2$ as an example)} \label{eqlSubNet}
\end{figure}

The same expression may have multiple equivalent subnetworks represented in PruneSymNet. Which subnetwork is obtained by pruning algorithm is not controllable, and sometimes it is not necessarily the simplest subnetwork. For example, the simplest subnetwork of the expression $y={x_1}^2+{x_2}^2$ is shown in Figure \ref{eqlSubNet}(a). But the subnetwork we obtained during one pruning is shown in Figure \ref{eqlSubNet}(b) that was written as Equation (\ref{equ:eql1}), where $C$ represents the coefficients. Substitute the true value of $C$ in Equation (\ref{equ:eql1}) to obtain Equation (\ref{equ:eql2}), and simplify Equation (\ref{equ:eql2}) to obtain Equation (\ref{equ:eql3}) that is the optimal solution.
\begin{equation}\label{equ:eql1}
\begin{gathered}
y=(x_2\times C\times x_2\times C\times C+x_2\times C\times x_2\times C\times C)\times C+(x_2\times C\times x_2\times C\times C\\-
(x_1\times C\times x_1\times C\times C-x_1\times C\times x_1\times C\times C)\times C)\times C
\end{gathered}
\end{equation}
\begin{equation}\label{equ:eql2}
\begin{gathered}
y=(x_2\times 1.0928603410720825\times x_2\times 1.0928603410720825\times 0.5980429649353027+\\x_2\times 1.0928603410720825\times x_2\times 1.0928603410720825\times 0.5980429649353027)\times \\ 0.32933270931243896+(x_2\times 1.0928603410720825\times x_2\times 1.0928603410720825\times \\ 0.7715291976928711-(x_1\times 0.6822513341903687\times x_1\times 0.6822513341903687\times \\ 1.517747402191162-x_1\times 0.6309414505958557\times x_1\times 0.6309414505958557\times \\ (-2.3672585487365723))\times (-1.0532234907150269))\times 0.5690650343894958
\end{gathered}
\end{equation}
\begin{equation}\label{equ:eql3}
y = 0.98823441142276889 \times {x_1}^2{\text{ }} + {\text{ }}0.99484115749571473 \times {x_2}^2
\end{equation}
\subsection{Limitations of the proposed algorithm}
There are two main limitations of the proposed algorithm. One is that it is not good for non convex expressions, and the other is that the algorithm is sensitive to the range of variable values.

 It is easy to get into saddle points for solving non-convex expressions(such as trigonometric functions), since the pruning algorithm is based on gradient descent method, which will affect the solving performance. It can be seen from the solution of trigonometric functions in Table 1-6 in supplementary materials that the optimal solution cannot be obtained for some simple trigonometric functions.

\begin{figure}[!htbp]
	\centering
	\includegraphics[width=0.2\textwidth]{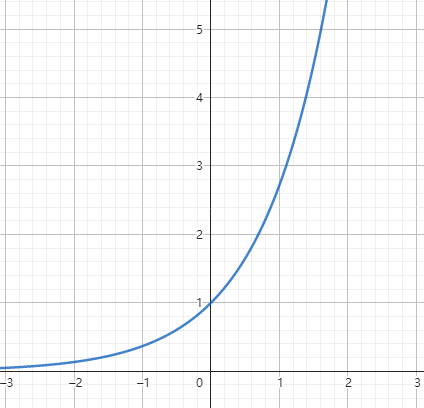}
	\caption{The curve of $y=e^x$} \label{exp}
\end{figure}

If the range of variable values is too small, the solution result may be incorrect, because the sampling points are not sufficient to express the characteristics of the original expression. For example, for the curve corresponding to $y=e^x$, as shown in Figure \ref{exp}, if the variable value range is [0,1], the curve is close to a straight line, and it is easy to obtain the result of $y=kx+b$. If the value range is expanded to [-4,4], it is easy to obtain the optimal solution.
%\iffalse
%\fi
\section{Conclusion}\label{sec:conclusion}
In this study, a neural network PruneSymNet is proposed for symbolic regression, which can represent any symbolic expression. The whole network is differentiable and can be trained using gradient descent method. The problem of gradient explosion in the training process is solved so that the network can be trained stably even with division operators. In order to extract concise and interpretable expressions, we proposed a greedy pruning algorithm based on gradient descent to prune the network into a minimalist subnetwork. 
%In order to solve the problem that greedy pruning can not get the optimal solution,
we introduced beam search to get multiple candidate expressions to improve the greedy pruning algorithm. Randomness was introduced to increase exploration in order to solve the problem of local optimal solution. The results showed that the proposed algorithm is superior to other popular algorithms.
%The problem that greedy pruning can not get the optimal solution is alleviated by introducing multiple candidate expressions in beam search. By introducing randomness to increase exploration, we solved the problem of local optimal solutions. The results showed the superiority of the proposed algorithm.

\section{Supplementary Material}
\iffalse
Authors may wish to optionally include extra information (complete proofs, additional experiments and plots) in the appendix. All such materials should be part of the supplemental material (submitted separately) and should NOT be included in the main submission.

The public datasets used in the experiment (Koza, Korns \cite{gen}, ODE\cite{zongshu}, Livermore\cite {dso}, AIFeyman2\cite {AIFeyman}, AIFeyman3 \cite{AIFeyman}) are shown in Table \ref{LA2-8} - \ref{LA2-13}, which indicates whether the testing algorithm has found the optimal solution.
\fi
The public datasets(Koza, Korns, ODE, Livermore, AIFeyman2, AIFeyman3) used in the experiment are shown in Table \ref{LA_1}--\ref{LA_6}.  Every algorithm was independently tested on the test set for five times in order to evaluate the performance of the algorithm in obtaining the optimal solution. The number of times each algorithm obtained the optimal solution for each expression was also listed in Table \ref{LA_1}--\ref{LA_6}. It can be seen from the results that PruneSymNet obtained the most optimal solution.
All source code can be found in an anonymous website https://github.com/wumin86/PruneSymNet.
%, which indicates whether the testing algorithm has found the optimal solution.
\begin{table}[!htbp]
	\centering
	\caption{The Koza dataset and the number of times the optimal solution was obtained}\label{LA_1}
	\begin{tabular}{ccccc}
		\hline
		expression&EQL&GP&DSR&PruneSymNet\\
		\hline
		${x^4} + {x^3} + {x^2} + x$ & 0 &  0 &\bf  5 &\bf  5 \\
		\hline
		${x^5} - 2{x^3} + x$ & 0 &  0 &\bf  5 &\bf  5 \\
		\hline
		${x^6} - 2{x^4} + {x^2}$ & 0 &  0 &  0 &\bf  5 \\
		\hline
		${x^3} + {x^2} + x$ & 0 &  2 &\bf  5 &\bf  5 \\
		\hline
		${x^5} + {x^4} + {x^3} + {x^2} + x$ & 0 &  0 &\bf  5 &\bf  5 \\
		\hline
		${x^6} + {x^5} + {x^4} + {x^3} + {x^2} + x$ & 0 &  0 &\bf  5 &\bf  5 \\
		\hline
		$\sin ({x^2})\cos (x) - 1$ & 0 &  0 &  0 &  0 \\
		\hline
		$\sin \left( x \right) + {\text{sin}}\left( {x + {x^2}} \right)$ & 0 &  0 &\bf  5 &  0 \\
		\hline
		$\log (x + 1) + \log ({x^2} + 1)$ & 0 &  0 &  0 &  0 \\
		\hline
		$\sin \left( x \right) + {\text{sin}}\left( {{y^2}} \right)$ & 0 &  0 &\bf  5 &  0 \\
		\hline
		$2\sin \left( x \right)cos\left( y \right)$ & 0 &  0 &\bf  5 &  0 \\
		\hline
		${x^4} - {x^2} + \frac{1}{2}{y^2} - y$ & 0 &  0 &  0 &\bf  1 \\
		\hline
		$3.39{x^3} + 2.12{x^2} + 1.78x$ & 0 &  0 &  0 &\bf  5 \\
		\hline
		$0.48{x^4} + 3.39{x^3} + 2.12{x^2} + 1.78x$ & 0 &  0 &  0 &\bf  5 \\
		\hline
		$\sin ({x^2})\cos (x) - 0.75$ & 0 &  0 &  0 &  0 \\
		\hline
		$\sin (1.5x) + \sin (0.5{y^2})$ & 0 &  0 &  0 &  0 \\
		\hline
		Total & 0 &  2 &  40 &  \bf 41 \\
		\hline
		
	\end{tabular}
\end{table}

\begin{table}[!htbp]
	\centering
	\caption{The Korns dataset and the number of times the optimal solution was obtained}\label{LA_2}
	\begin{tabular}{ccccc}
		\hline
		expression&EQL&GP&DSR&PruneSymNet\\
		\hline
		$1.57 + 24.3x$ & 0 &  0 &  0 &\bf  5 \\
		\hline
		$ -2.3 + 0.13sin(x)$ & 0 &  0 &  0 &\bf  2 \\
		\hline
		$3 + 2.13\log (x)$ & 0 &  0 &  0 &\bf  5 \\
		\hline
		$213.80940889(1 - {e^{ - {\text{0}}{\text{.54723748542}}x}})$ & 0 &  0 &  0 &\bf  5 \\
		\hline
		$6.87 + 11\cos (7.23{x^3})$ & 0 &  0 &  0 &  0 \\
		\hline
		$2 - 2.1\cos (9.8x)\sin (1.3y)$ & 0 &  0 &  0 &  0 \\
		\hline
		$0.23 + 14.2*\frac{{(x + y)}}{{3z}}$ & 0 &  0 &  0 &\bf  3 \\
		\hline
		Total & 0 &  0 &  0 &  \bf 20 \\
		\hline

	\end{tabular}
\end{table}

\begin{table}[!htbp]
	\centering
	\caption{The ODE dataset and the number of times the optimal solution was obtained}\label{LA_3}
	\begin{tabular}{ccccc}
		\hline
		expression&EQL&GP&DSR&PruneSymNet\\
		\hline
		$20 - x - \frac{{xy}}{{1 + 0.5{x^2}}}$ & 0 &  0 &  0 &  0 \\
		\hline
		$10 - \frac{{xy}}{{1 + 0.5{x^2}}}$ & 0 &  0 &  0 &  0 \\
		\hline
		$0.5\sin (x - y) - \sin (x)$ & 0 &  0 &  0 &  0 \\
		\hline
		$ - 0.05{x^2} - \sin (y)$ & 0 &  0 &  0 &\bf  1 \\
		\hline
		$x - \frac{{\cos (y)}}{x}$ & 0 &  0 &\bf  5 &  0 \\
		\hline
		$3x - 2xy - {x^2}$ & 0 &  2 &\bf  5 &\bf  5 \\
		\hline
		$2y - xy - {y^2}$ & 0 &\bf  5 &\bf  5 &\bf  5 \\
		\hline
		$x(4 - x - \frac{y}{{1 + x}})$ & 0 &  0 &  0 &  0 \\
		\hline
		$y(\frac{x}{{1 + x}} - 0.075y)$ & 0 &  0 &  0 &  0 \\
		\hline
		$({\cos ^2}(x) + 0.1{\sin ^2}(x))\sin (y)$ & 0 &  0 &  0 &  0 \\
		\hline
		$10(y - \frac{1}{3}({x^3} - x))$ & 0 &  0 &  0 &\bf  4 \\
		\hline
		$ - \frac{1}{{10}}x$ & 0 &  1 &  0 &\bf  5 \\
		\hline
		Total & 0 &  8 &  15 & \bf 20 \\
		\hline

	\end{tabular}
\end{table}

\begin{table}[!htbp]
	\centering
	\caption{The Livermore dataset and the number of times the optimal solution was obtained}\label{LA_4}
	\begin{tabular}{ccccc}
		\hline
		expression&EQL&GP&DSR&PruneSymNet\\
		
		\hline
		$\frac{1}{3} + x + \sin ({x^2})$ & 0 &  0 &  0 &  0 \\
		\hline
		$\sin ({x^2})\cos (x) - 2$ & 0 &  0 &\bf  5 &  0 \\
		\hline
		$\sin ({x^3})\cos ({x^2}) - 1$ & 0 &  0 &  0 &  0 \\
		\hline
		$\log (x + 1) + \log ({x^2} + 1) + \log (x)$ & 0 &  0 &  0 &  0 \\
		\hline
		${x^4} - {x^3} + {x^2} - x$ & 0 &  3 &\bf  5 &\bf  5 \\
		\hline
		$4{x^4} + 3{x^3} + 2{x^2} + x$ & 0 &  0 &\bf  5 &\bf  5 \\
		\hline
		${x^9} + {x^8} + {x^7} + {x^6} + {x^5} + {x^4} + {x^3} + {x^2} + x$ & 0 &  0 &  0 &\bf  2 \\
		\hline
		$6\sin (x)\cos (y)$ & 0 &  0 &  0 &  0 \\
		\hline
		$\frac{{{x^5}}}{{{y^3}}}$ & 0 &  0 &  0 &\bf  1 \\
		\hline
		${x^3} + {x^2} + x + \sin (x) + \sin ({x^2})$ & 0 &  0 &  0 &  0 \\
		\hline
		$4\sin (x)\cos (y)$ & 0 &  0 &\bf  5 &  0 \\
		\hline
		$\sin ({x^2})\cos (y) - 5$ & 0 &  0 &  0 &  0 \\
		\hline
		${x^5} + {x^4} + {x^2} + x$ & 0 &  2 &\bf  5 &\bf  5 \\
		\hline
		${e^{ - {x^2}}}$ & 0 &  0 &\bf  5 &\bf  5 \\
		\hline
		${x^8} + {x^7} + {x^6} + {x^5} + {x^4} + {x^3} + {x^2} + x$ & 0 &  0 &  0 &\bf  5 \\
		\hline
		${e^{ - 0.5{x^2}}}$ & 0 &  0 &  0 &\bf  5 \\
		\hline
		Total & 0 &  5 &  30 & \bf 33 \\
		\hline
		
	\end{tabular}
\end{table}

\begin{table}[!htbp]
	\centering
	\caption{The AIFeynman2 dataset and the number of times the optimal solution was obtained}\label{LA_5}
	\begin{tabular}{ccccc}
		\hline
		expression&EQL&GP&DSR&PruneSymNet\\
		
		\hline
		$\frac{{\sqrt 2 {e^{ - \frac{{{x^2}}}{2}}}}}{{2\sqrt \pi  }}$ & 0 &  0 &  0 &\bf  5 \\
		\hline
		$\frac{{\sqrt 2 {e^{ - \frac{{{y^2}}}{{2{x^2}}}}}}}{{2\sqrt \pi  x}}$ & 0 &  0 &  0 &  0 \\
		\hline
		$xy$ & 0 &\bf 5 &\bf  5 &\bf  5 \\
		\hline
		$\frac{{{x^2}y}}{2}$ & 0 &  1 &  0 &\bf 5 \\
		\hline
		$\frac{x}{y}$ & 0 &\bf 5 &\bf  5 &\bf 5 \\
		\hline
		$\frac{{xy}}{{2\pi }}$ & 0 &  0 &  0 &\bf 5 \\
		\hline
		$\frac{{3xy}}{2}$ & 0 &  0 &\bf 5 &\bf 5 \\
		\hline
		$\frac{x}{{4\pi {y^2}}}$ & 0 &  0 &  0 &\bf 2 \\
		\hline
		$\frac{{x{y^2}}}{2}$ & 0 &  1 &  0 &\bf 5 \\
		\hline
		$\frac{{xy}}{{ - \frac{{xy}}{3} + 1}} + 1$ & 0 &  0 &  0 &  0 \\
		\hline
		$x{y^2}$ & 0 &\bf 5 &\bf 5 &\bf 5 \\
		\hline
		$\frac{x}{{2y + 2}}$ & 0 &  0 &  0 &  0 \\
		\hline
		Total & 0 & 17 &  20 &  \bf 42 \\
		\hline
	\end{tabular}
\end{table}

\begin{table}[!htbp]
	\centering
	\caption{The AIFeynman3 dataset and the number of times the optimal solution was obtained}\label{LA_6}
	\begin{tabular}{ccccc}
		\hline
		expression&EQL&GP&DSR&PruneSymNet\\
		\hline
		$\frac{{\sqrt 2 {e^{ - \frac{{{{(y - z)}^2}}}{{2{x^2}}}}}}}{{2\sqrt \pi  x}}$ & 0 &  0 &  0 &  0 \\
		\hline
		$\frac{x}{{4\pi y{z^2}}}$ & 0 &  0 &  0 &  0 \\
		\hline
		$xyz$ & 0 &\bf 5 &\bf 5 &\bf 5 \\
		\hline
		$\frac{{y + z}}{{1 + \frac{{yz}}{{{x^2}}}}}$ & 0 &  0 &  0 &  0 \\
		\hline
		$xy\sin (z)$ & 0 &  0 &\bf 5 &  0 \\
		\hline
		$\frac{1}{{\frac{z}{y} + \frac{1}{x}}}$ & 0 &  0 &  0 &  0 \\
		\hline
		$\frac{{x{{\sin }^2}(\frac{{yz}}{2})}}{{{{\sin }^2}(\frac{y}{2})}}$ & 0 &  0 &  0 &  0 \\
		\hline
		$\frac{z}{{1 - \frac{y}{x}}}$ & 0 &  2 &\bf 5 &  0 \\
		\hline
		$\frac{{yz}}{{x - 1}}$ & 0 &  0 &  0 &  0 \\
		\hline
		$\frac{x}{{4\pi yz}}$ & 0 &  1 &  0 &\bf 3 \\
		\hline
		$\frac{{3{x^2}}}{{20\pi yz}}$ & 0 &  0 &  0 &\bf 1 \\
		\hline
		$\frac{x}{{y(z + 1)}}$ & 0 &\bf 1 &\bf 1 &  0 \\
		\hline
		$ - xy\cos (z)$ & 0 &  0 &  0 &\bf 1 \\
		\hline
		$xy{z^2}$ & 0 &  3 &\bf 5 &  4 \\
		\hline
		$\frac{{xy}}{{2\pi z}}$ & 0 &  1 &  0 &\bf 4 \\
		\hline
		$\frac{{xyz}}{2}$ & 0 &  0 &  0 &\bf 5 \\
		\hline
		$\frac{{xy}}{{4\pi z}}$ & 0 &  0 &  0 &\bf 4 \\
		\hline
		$xy(z + 1)$ & 0 &  4 &\bf 5 &  3 \\
		\hline
		$\frac{{4\pi xy}}{z}$ & 0 &  0 &  0 &\bf 3 \\
		\hline
		${\sin ^2}(\frac{{2\pi xy}}{z})$ & 0 &  0 &  0 &  0 \\
		\hline
		$2x(1 - \cos (yz))$ & 0 &  0 &  0 &  0 \\
		\hline
		$\frac{{{x^2}}}{{8{\pi ^2}y{z^2}}}$ & 0 &  0 &  0 &  0 \\
		\hline
		$\frac{{2\pi x}}{{yz}}$ & 0 &  0 &  0 &  1 \\
		\hline
		$x(y\cos (z) + 1)$ & 0 &  0 &\bf 5 &  0 \\
		\hline
		Total & 0 &  17 & 31 &\bf 34 \\
		\hline	
	\end{tabular}
\end{table}

\iffalse
\section*{References}

References follow the acknowledgments in the camera-ready paper. Use unnumbered first-level heading for
the references. Any choice of citation style is acceptable as long as you are
consistent. It is permissible to reduce the font size to \verb+small+ (9 point)
when listing the references.
Note that the Reference section does not count towards the page limit.
\medskip

{
\small

[1] Alexander, J.A.\ \& Mozer, M.C.\ (1995) Template-based algorithms for
connectionist rule extraction. In G.\ Tesauro, D.S.\ Touretzky and T.K.\ Leen
(eds.), {\it Advances in Neural Information Processing Systems 7},
pp.\ 609--616. Cambridge, MA: MIT Press.

[2] Bower, J.M.\ \& Beeman, D.\ (1995) {\it The Book of GENESIS: Exploring
  Realistic Neural Models with the GEneral NEural SImulation System.}  New York:
TELOS/Springer--Verlag.

[3] Hasselmo, M.E., Schnell, E.\ \& Barkai, E.\ (1995) Dynamics of learning and
recall at excitatory recurrent synapses and cholinergic modulation in rat
hippocampal region CA3. {\it Journal of Neuroscience} {\bf 15}(7):5249-5262.
}
\fi

\bibliographystyle{plainnat}
\bibliography{PruneSymNet.bib}

%\section{Supplementary Material}
%%%%%%%%%%%%%%%%%%%%%%%%%%%%%%%%%%%%%%%%%%%%%%%%%%%%%%%%%%%%

\end{document}